\documentclass[
]{ceurart}

\usepackage{listings}
\usepackage{amssymb}
\usepackage{makecell}
\usepackage{amsthm}
\usepackage{amsmath,amsfonts}

\newtheorem{definition}{Definition}
\newtheorem{problem}{Problem}
\newtheorem{lemma}{Lemma}
\usepackage{amsmath,amsfonts}
\usepackage{dsfont}
\usepackage{algorithm}
\usepackage{algorithmicx}
\usepackage{algpseudocode}
\usepackage{subcaption}
\usepackage{wrapfig}
\usepackage{multirow}

\newcommand{\refdef}[1]{Definition~\ref{#1}}    
\newcommand{\refeq}[1]{Eq.~\eqref{#1}}

\newcommand{\dimsym}{d}

\newcommand{\RN}{\mathbb{R}}
\DeclareMathOperator*{\argmin}{\arg\min}    
\DeclareMathOperator*{\loss}{{\ell}}
\DeclareMathOperator*{\dist}{{d}}

\newcommand{\set}[1]{\mathcal{#1}}
\newcommand{\pnorm}[1]{\lVert{#1}\rVert}    
\DeclareMathOperator*{\regularization}{\ensuremath{{\theta}}}

\newcommand{\classifier}{\ensuremath{h}}
\newcommand{\x}{\ensuremath{\vec{x}}}
\newcommand{\xorig}{\ensuremath{\vec{x}_\text{orig}}}
\newcommand{\y}{\ensuremath{\vec{y}}}
  
\newcommand{\xcf}{\ensuremath{\vec{x}_{\text{cf}}}}   
\newcommand{\deltacf}{\ensuremath{\vec{\delta}_{\text{cf}}}}    
\newcommand{\ycf}{\ensuremath{y_{\text{cf}}}}

\newcommand{\setX}{\ensuremath{\set{X}}}  
\newcommand{\setY}{\ensuremath{\set{Y}}}  

\newcommand{\myCF}[2]{\ensuremath{\text{CF}_{#2}(#1)}}
\begin{document}

\copyrightyear{2024}
\copyrightclause{Copyright for this paper by its authors.
  Use permitted under Creative Commons License Attribution 4.0
  International (CC BY 4.0).}

\conference{Late-breaking works, 2nd World Conference on eXplainable Artificial Intelligence}

\title{A Two-Stage Algorithm for Cost-Efficient Multi-instance Counterfactual Explanations}

%\tnotetext[1]{You can use this document as the template for preparing your
%  publication. We recommend using the latest version of the ceurart style.}

\author[1,2]{Andr\'e Artelt}[%
orcid=0000-0002-2426-3126,
email=aartelt@techfak.uni-bielefeld.de,
%url=https://andreartelt.github.io/,
]
\address[1]{Faculty of Technology, Bielefeld University, Inspiration 1, 33615 Bielefeld, Germany}
\address[2]{University of Cyprus, Panepistimiou 1, 2109 Aglantzia, Nicosia, Cyprus}

\author[3]{Andreas Gregoriades}[%
orcid=0000-0002-7422-1514,
email=andreas.gregoriades@cut.ac.cy,
]
\address[3]{Department of Management, Entrepreneurship and Digital Business, Cyprus University of Technology, 30 Arch. Kyprianos Str., 3036 Limassol, Cyprus}

\begin{abstract}
Counterfactual explanations constitute among the most popular methods for analyzing black-box systems since they can recommend cost-efficient and actionable changes to the input of a system to obtain the desired system output. While most of the existing counterfactual methods explain a single instance, several real-world problems, such as customer satisfaction, require the identification of a single counterfactual that can satisfy multiple instances (e.g. customers) simultaneously.
To address this limitation, in this work, we propose a flexible two-stage algorithm for finding groups of instances and computing cost-efficient multi-instance counterfactual explanations. The paper presents the algorithm and its performance against popular alternatives through a comparative evaluation.
\end{abstract}


\maketitle

\section{Introduction}
Recently an increasing number of Artificial Intelligence (AI) systems have been applied to important problems, such as medical image classification ~\cite{goyal2020artificial} and many more.
Although these systems show impressive performance when used on experimental data, they are still imperfect when applied to real-world problems, and in some cases can cause harm to humans due to biases embedded in their logic ~\cite{ferrer2021bias}.
Therefore, transparency of such systems is of paramount importance, since it assists in understanding their logic and thus allows decision-makers to decide where and how it is safe to deploy them~\cite{larsson2020transparency}. The importance of transparency is also stressed at EU level, with recent regulations such as the AI act~\cite{euAiAct21} making explicit reference to the need for explainability.
The field of explainability is not new and focuses on answering "why" a system behaves in a certain way. Recently the term eXplainable AI (XAI)~\cite{dwivedi2023explainable} has been coined, which boosted the popularity of the field and led to the introduction of many different XAI methods in different domains  ~\cite{dwivedi2023explainable,rawal2021recent}. One of the most popular types of explanation methods are counterfactual explanations~\cite{CounterfactualWachter}, which mimic the way humans seek  explanations~\cite{CounterfactualsHumanReasoning}.
By definition, a counterfactual explanation provides actionable recommendations on how to change a predictive system's output in some desired way -- e.g. how to change a rejected loan application into an accepted one.
In many business problems where AI systems are deployed such as customer repurchase prediction (how to make customers buy again from a firm), and employee attrition (how to prevent employees from leaving the organisation), the decision maker is not only interested in explaining a single instance of the predictive system but a group of instances -- e.g. how to prevent many employees from quitting, instead of only one.
%Thus, there is a need to find a single explanation that can satisfy (i.e. holds for) a group of instances.
To address such use cases, the concept of multi-instance counterfactual explanations has been recently introduced~\cite{kanamori2022counterfactual,warren2023explaining}. Here, the aim is to identify a single explanation of how to change the system's output for a group of instances simultaneously. Because of the novelty of this concept, many issues still exist -- in particular, how to identify groups of instances for which cost-efficient multi-instance counterfactual explanations can be computed.
\textbf{Our contributions:} In this work, we formalize and investigate the problem of finding cost-efficient counterfactual explanations for groups of instances (multi-instance). Based on our formal analysis, we propose a model (data-agnostic) two-stage algorithm for computing such multi-instance counterfactual explanations.

\section{Foundations}\label{sec:foundations}
A counterfactual explanation (often just called counterfactual) proposes cost-effective and actionable changes to the features of a given input instance of a model such that its prediction changes to the desired output.
Because counterfactuals mimic human explanations~\cite{CounterfactualsHumanReasoning}, they constitute among the most popular explanation methods and a favorable choice in many practical problems~\cite{CounterfactualReviewChallenges}.
The computation of counterfactual explanations involves the consideration of two important aspects~\cite{CounterfactualWachter,keane2020good}: 1) the contrasting property, which requires that the stated changes indeed alter the output of the system, and 2) the cost of the counterfactual, which defines the difficulty and effort it takes to realise the explanation (i.e. recommendations) in the real world.
\begin{definition}[Counterfactual Explanation]\label{def:counterfactual}
Assume a prediction function $\classifier:\setX \to \setY$ is given. Computing a counterfactual explanation $\deltacf \in \setX$ for a given instance $\xorig \in \setX$ is phrased as the following optimization problem:
%\begin{equation}\label{eq:counterfactualoptproblem}
$\underset{\deltacf \,\in\, \setX}{\arg\min}\; \loss\big(\classifier(\xorig \oplus \deltacf), \ycf\big) + C \cdot \regularization(\deltacf)$
%\end{equation}
%
where $\loss(\cdot)$ denotes a loss function that penalizes deviation of the output $\classifier(\xorig \oplus \deltacf)$ from the requested output $\ycf$, $\regularization(\cdot)$ states the cost of $\deltacf$, and $C>0$ denotes the regularization strength.
\end{definition}
The symbol $\oplus$ denotes the application/execution of the counterfactual $\deltacf$ to the original instance $\xorig$ -- i.e. for $\setX=\RN^\dimsym$ this reduces to $(\xcf)_i = (\xorig)_i + {(\deltacf)}_i$, and to $(\xcf)_i = {(\deltacf)}_i$ in the case of categorical features.
Also note that the cost of the counterfactual, here modeled by $\regularization(\cdot)$, is highly domain and often use-case specific~\cite{guidotti2022counterfactual}, with  $p$-norm  being the default.
Furthermore, there exist additional relevant aspects such as plausibility~\cite{poyiadzi2020face},
robustness~\cite{artelt2021evaluating,slack2021counterfactual}, fairness~\cite{artelt2023ijcai}, etc.
The basic formalization~\refdef{def:counterfactual}, however, is still very popular and widely used in practice~\cite{CounterfactualReviewChallenges,guidotti2022counterfactual}.

Most existing methods do not address the case where we have to assign the same actions to multiple instances simultaneously~\cite{kanamori2022counterfactual} -- i.e. explain more than a single instance within a single counterfactual~\cite{ley2023globe,kanamori2022counterfactual}.
In contrast to~\refdef{def:counterfactual}, a multi-instance counterfactual states a single change $\deltacf$ that alters the output of a classifier $\classifier:\setX \to \setY$ for many instances $\x_i\in\setX$ simultaneously.
While multi-instance counterfactuals are formalized slightly differently in the literature  ~\cite{warren2023explaining,ley2023globe,kanamori2022counterfactual,ArteltG23}, related work agrees that the same two properties as in~\refdef{def:counterfactual} must be considered: 1) The cost of the explanation $\deltacf$ 2) an extension of the contrasting property from~\refdef{def:counterfactual}: the explanation $\delta$ should be valid for all/many instances in a given set of instances $\set{D}$.
In this work, we formalize a multi-instance counterfactuals explanation as follows: 
%a multi-objective optimization problem as stated in~\refdef{def:multiinstcf}.
\begin{definition}[Multi-instance Counterfactual Explanation]\label{def:multiinstcf}
Let $\classifier:\setX\to\setY$ denote a prediction function, and let $\set{D}$ be a set of labeled instances with the same prediction $y\in\setY$ under $\classifier(\cdot)$ -- i.e. $\classifier(\x_i)=y \quad\forall\x_i\in\set{D}$. 
We call all pareto-optimal solutions $\deltacf$ to the following multi-objective optimization problem "multi-instance counterfactuals":

%\begin{equation}\label{eq:setconsistentcf:relaxed}
$\underset{\deltacf\,\in\,\setX}{\min}\; \big(\regularization(\deltacf) \;,\; \loss(\classifier(\x_1 \oplus \deltacf), \ycf), \dots, \loss(\classifier(\x_{|\set{D}|} \oplus \deltacf), \ycf)\big)$
%\end{equation}
%where $\regularization(\cdot)$ denotes the cost of the counterfactual, and $\loss(\cdot)$ denotes a suitable loss function penalizing deviations from the requested outcome $\ycf$ -- suitable loss functions might be the mean-squared error or cross-entropy loss, while the cost $\regularization(\cdot)$ might be implemented by a $p$-norm.
\end{definition}
Note that the main difference to ~\refdef{def:counterfactual} is that the contrasting property leads to multiple objects (i.e. one objective for each instance in $\set{D}$) -- i.e. the change $\deltacf$ must be valid for all (or as many as possible) instances in $\set{D}$.

\paragraph{Related Work}
One of the earliest works that addresses this problem proposes a counterfactual explanation tree~\cite{kanamori2022counterfactual}, which assigns counterfactuals to the instances at the leaves of a decision tree, derived from $\set{D}$ -- i.e. each leaf is interpreted as a group. This method groups instances and computes multi-instance counterfactuals in a single step. While this might be beneficial in some scenarios, it also constitutes a limitation since the user cannot customize the groupings and also lacks any formal guarantees due to the use of a heuristic (local search) in the implementation.
In general, a large part of existing work for multi-instance counterfactuals can be interpreted as summarizing or aggregating individual counterfactual explanations~\cite{warren2023explaining,ley2023globe,DBLP:conf/nips/RawalL20,plumb2020explaining}.
For instance, in~\cite{warren2023explaining}, multi-instance counterfactuals are generated by first computing individual counterfactuals and then applying a sampling strategy to select the one that satisfies most instances from a given set,  for which a multi-instance counterfactual is requested.
In previous work~\cite{ArteltG23}, multi-instance counterfactuals are implemented utilizing convex programming for linear classifiers only.
However, those methods assume that a grouping is already given and also often suffer from poor performance (e.g. low coverage and correctness).
A related branch of research is counterfactual robustness with respect to input changes~\cite{leofante2024promoting,dominguez2022adversarial,artelt2021evaluating}.
Robust counterfactuals~\cite{leofante2024promoting} should not change for similar instances -- i.e. those robust counterfactuals would constitute multi-instance counterfactuals for their local neighborhood in data space. However, if instances are too different from each other, robust counterfactuals do not provide a solution to the multi-instance counterfactual explanation problem.

\section{A Two-Stage Algorithm for Multi-instance Counterfactuals}\label{sec:method}
As stated in~\refdef{def:multiinstcf}, a multi-instance counterfactual states changes $\deltacf$ that are valid for a set of instances $\set{D}$. While in some scenarios, the $\set{D}$ might be given a priori, in other scenarios it might be more flexible and require finding groups along with cost-efficient multi-instance counterfactuals. For instance, business owners might be interested in identifying groups of customers along with recommendations on how to improve their repurchase intentions.
In these cases, it is important to identify large groups of instances for which cost-efficient multi-instance counterfactuals (\refdef{def:multiinstcf}) can be computed. We formalize this as a multi-objective optimization problem as stated in Problem~\ref{prob:grouping}.
\begin{problem}\label{prob:grouping}
    For a classifier $\classifier:\setX\to\setY$ and a set of instances $\set{D}\subset\setX^n$ with $\classifier(\x_i)=y \;\forall\x_i\in\set{D}, y\in\setY$, we are looking for $N$ partitions $\set{G}_i$ of the instances such that cost-efficient multi-instance counterfactuals (\refdef{def:multiinstcf}) exists:
    \begin{subequations}
    \begin{align}
            &\min N  \quad \max \big(|\set{G}_1|, \dots, |\set{G}_N|\big) \quad \text{s.t. } \underset{i}{\bigcup}\, \set{G}_i = \set{D}, \quad \set{G}_i \cap \set{G}_j = \emptyset\; \forall i\neq j\label{eq:prob:grouping:size_clusters} \\
            &\min \big(\regularization({\deltacf}_1), \dots, \regularization({\deltacf}_N)\big) \quad \min \big(\loss(\classifier(\x_j \oplus \deltacf), \ycf) \;\x_j\in\set{G}_i, \dots,\set{G}_N\big) \label{eq:prob:grouping:acc_clusters}
        \end{align}
    \end{subequations}
\end{problem}

In this work we study Problem~\ref{prob:grouping} and propose the following process for computing multi-instance counterfactuals: Stage 1) Finding a grouping of instances and then Stage 2) Computing multi-instance counterfactual explanations for each of those groups -- by this, we aim to reduce the effect of outliers on the cost of the final multi-instance counterfactuals.

\paragraph{Stage 1- Grouping of Instances.}
For this task, a naive approach would have been to group the instances based on their spatial similarity/distances -- e.g. by using a clustering method such as k-means. However, because counterfactuals are known not to be robust with respect to large changes in the input~\cite{artelt2021evaluating}, this approach is likely to fail. Furthermore, such an approach does not take into account any knowledge about the cost $\regularization(\cdot)$, which is necessary to compute cost-efficient counterfactuals -- we empirically confirm this in the experiments in Section~\ref{sec:experiments}.
\begin{lemma}\label{lemma:1}
    Assume a linear binary classifier $\classifier:\RN^\dimsym \to \{0,1\}$ and $\regularization(\cdot)=\pnorm{\cdot}_p$.
    Furthermore, for a given set of instances $\x_i\in\RN^\dimsym$ we denote their counterfactual explanation (\refdef{def:counterfactual}) as ${\deltacf}_i$.
    If $\forall\, i\neq j:\; {\deltacf}_i^\top {\deltacf}_j = \pnorm{{\deltacf}_i}_2 \cdot \pnorm{{\deltacf}_j}_2$, then the  cost $\regularization(\cdot)$ of the multi-instance counterfactual $\deltacf$ (\refdef{def:multiinstcf}) is given as
    %\begin{equation}
        $\regularization(\deltacf) = \underset{i}{\max}\, \regularization({\deltacf}_i)$
    %\end{equation}
\end{lemma}
%\begin{proof}
%Sketch: ${\deltacf}_i^\top {\deltacf}_j = \pnorm{{\deltacf}_i}_2 \cdot \pnorm{{\deltacf}_j}_2 \;\forall i\neq j$ implies that $\exists \alpha_j\in\RN:\,{\deltacf}_j=\alpha_j\cdot {\deltacf}_i \;\forall j$. Monotonicty of $\classifier(\cdot)$ implies $\exists \alpha\in\RN:\,{\deltacf}_j=\alpha\cdot {\deltacf}_i\;\forall j$. The statement follows from selecting $\alpha$ and ${\deltacf}_i$
%\end{proof}
Lemma~\ref{lemma:1} states that if the individual counterfactuals all have the same direction, then a multi-instance counterfactual not only exists but we can also state a tight upper bound on its cost.
Although Lemma~\ref{lemma:1} is stated for a linear classifier, it can also be applied to arbitrary classifiers that can be approximated locally by a linear classifier.
This suggests that groups of instances where the individual counterfactuals (\refdef{def:counterfactual}) point in similar directions are good candidates for which cost-efficient multi-instance counterfactuals (\refdef{def:multiinstcf}) might exist.
We, therefore, propose to 1) compute single counterfactuals (\refdef{def:counterfactual}) for each instance, and then 2) cluster those into groups based on their direction (i.e. based on their cosine similarity) -- optionally, in addition, one could also cluster in a second step according to their amount of change.
In the remainder of this work, we limit ourselves to minimizing the number of changes -- i.e. we cluster only based on the direction of the individual counterfactuals.
The number of groups (i.e. clusters) might be given by the user or might be determined automatically, e.g. using the Elbow method. The complete procedure is described in Algorithm~\ref{algo:multicf:grouping}.
\begin{algorithm}[t!]
\caption{Grouping of Instances For Cost-Efficient Multi-instance Counterfactuals}\label{algo:multicf:grouping}
\textbf{Input:} Instances $\x_i$ with the same prediction $\classifier(\x_i)=\y$, counterfactual generation $\myCF{\cdot}{\classifier}$\\
\textbf{Output:} Grouping of instances
\begin{algorithmic}[1]
  \State $\{{\deltacf}_i = \myCF{\x_i}{\classifier}\}$ \Comment{Compute a counterfactual ${\deltacf}_i$ for each instance $\x_i$}
  \For{Different number of clusters}    \Comment{Optimize number of clusters if requested/needed}
      \State Cluster with $\dist({\deltacf}_i, {\deltacf}_j)=\frac{{\deltacf}_i^\top{\deltacf}_j}{\pnorm{{\deltacf}_i}_2 \pnorm{{\deltacf}_j}_2}$ \Comment{Cluster based on the directions of ${\deltacf}_i$}
      \State Sub-cluster with $\dist({\deltacf}_i, {\deltacf}_j)=\pnorm{\regularization({\deltacf}_i) - \regularization({\deltacf}_j)}_2$  \Comment{Cluster based on the cost $\regularization({\deltacf}_i)$}
  \EndFor
\end{algorithmic}
\end{algorithm}

\paragraph{Stage 2- Computing Multi-instance Counterfactuals}\label{sec:method:ea}
Given a group of instances, we can use any existing method from the literature for computing multi-instance counterfactuals (\refdef{def:multiinstcf}). However, because existing model/domain-agnostic methods are limited and often show sub-optimal performance with respect to correctness, we propose an evolutionary method for solving~\refdef{def:multiinstcf}. This not only constitutes a model/domain-agnostic method but also a very flexible solution since additional constraints can be easily introduced.
Our evolutionary method is an instance of the classic $(\mu+\lambda)$ genetic algorithm~\cite{reeves2010genetic} and can handle all types of variables.
In order to guarantee the feasibility of the final multi-instance counterfactual $\deltacf$ for the given problem domain, we construct the set of feasible changes for each feature of numerical variables as follows -- assuming non-negativity which can be achieved by adding a constant:
$l_i = \alpha_i - \underset{j}{\min}\{(\x_j)_i\} \text{ and } u_i = \beta_i - \underset{j}{\max}\{(\x_j)_i\}$
, where $\alpha_i$ and $\beta_i$ denote the maximum and minimum feasible value of the $i$-th feature, and the final set of feasible changes is then given as $[l_i, u_i]$.
These sets are used when computing mutations in our evolutionary algorithm of existing individuals during the optimization.
Furthermore, we merge the objectives in~\refeq{eq:prob:grouping:acc_clusters} into a single objective as follows: $\underset{\deltacf\,\in\,\setX}{\argmin}\; \regularization(\deltacf) + C\cdot\sum_{\x_i\,\in\,\set{D}}\loss(\classifier(\x_j \oplus \deltacf), \ycf)$
where the cost $\regularization(\deltacf)$ is defined as:
$
    \regularization(\deltacf) = \sum_i \psi((\deltacf)_i) \text{ where } \psi((\deltacf)_i) = |(\deltacf)_i| \; \text{or} 1 \; \text{if $i$-feature is categorical.}
$

\section{Experiments}\label{sec:experiments}
The following experiments are conducted to showcase the application and merits of the proposed method. All experiments and the proof of Lemma~\ref{lemma:1} are publicly available on GitHub\footnote{\url{https://github.com/andreArtelt/TwoStageMultiinstCFs}}.

%\paragraph{Benchmark datasets}
We consider two datasets that have been used in other work on multi-instance counterfactuals~\cite{ArteltG23,kanamori2022counterfactual}:
The \textit{IBM human resource attrition dataset}~\cite{IbmHrAnalyticsEmployee} (\textit{Attr.}) containing $35$ features for $1467$ unique employees.
The \textit{Law school data set}~\cite{wightman1998lsac} (\textit{Law}) containing $20798$ law school admission records, each described by $12$ attributes.

\paragraph{Setup.}
We empirically compare our proposed evolutionary algorithm (denoted by \emph{EA}) from Section~\ref{sec:method:ea}, against two methods~\cite{warren2023explaining,kanamori2022counterfactual}, which similarly to our method are also model and data-agnostic.
In this context, we evaluate two properties of the computed multi-instance counterfactual explanations, and the results are shown in Table~\ref{tab:exp_results:accuracy} and Table~\ref{tab:exp_results:cost}. The properties are:
1) The correctness, i.e. evaluating for how many samples the explanation is correct: $\frac{1}{|\set{D}|}\sum_i \mathds{1}\left(\classifier(\x_i \oplus \deltacf) = \ycf\right)$
--  Table~\ref{tab:exp_results:accuracy}.
2) The cost $\regularization(\cdot)$ expressed as the number of changed features, i.e. $\regularization(\deltacf) = \sum_i  \mathds{1}\left((\deltacf)_i \neq 0\right)$ -  Table~\ref{tab:exp_results:cost}.
All experiments were done using a five-fold cross-validation and we report the mean and variance of the results over all folds.
An XGBoost classifier is fitted to the training set and all negatively classified instances (i.e. $\classifier(\x_i)=0$) from the test-set define the set $\set{D}$ used by the multi-instance counterfactual method.
We compute a multi-instance counterfactual for the entire set of selected instances $\set{D}$, and also cluster (using DBSCAN) the set $\set{D}$ in two different ways: 1) Clustering based on the individual counterfactuals as proposed in Algorithm~\ref{algo:multicf:grouping} using the cosine-similarity, and for comparison 2) clustering based on the individual instances $\x_i$ using the Euclidean distance.

\paragraph{Results \& Discussion.}
From the results, we observe that our proposed method achieves excellent performance (with respect to correctness and cost) across all settings.
The method by Warren et al ~\cite{warren2023explaining} often struggles to find correct multi-instance counterfactuals (i.e. counterfactuals that cover as many as possible instances), also their method almost always produces multi-instance counterfactuals that use all available features and therefore have a higher cost if implemented in practice.
The method by Kanamori et al. ~\cite{kanamori2022counterfactual} often achieves a competitive performance and yields the most cost-efficient solutions while sacrificing correctness -- however, this method automatically creates additional sub-groups and is therefore difficult to compare to the other methods.
Furthermore, we observe that our proposed clustering of individual counterfactuals in  Algorithm~\ref{algo:multicf:grouping} often improves significantly the correctness and the cost of the computed multi-instance counterfactuals.
The results demonstrate the merits of our proposed two-stage algorithm.
\begin{table}[t]{}
\caption{Correctness (in percentage) of the generated multi-instance counterfactuals. We report the mean and variance (over all folds) rounded to two decimal points -- \emph{larger} numbers are better.}
\label{tab:exp_results:accuracy}
\centering
%\small
\begin{tabular}{|c|c||c||c|c|}
    \hline
     Data & Method & No Clustering $\uparrow$ & Clustering $\x_i$ $\uparrow$ & Clustering CFs $\uparrow$ \\
     \hline\hline
     \multirow{3}{*}{{Attr.}}
     & EA [Ours] & $0.98 \pm 0.0$ & $0.95 \pm 0.02$ & $1.0 \pm 0.0$ \\
     & Warren et al.~\cite{warren2023explaining} & $0.37 \pm 0.02$ & $0.34 \pm 0.04$ & $0.52 \pm 0.06$ \\
     & Kanamori et al.~\cite{kanamori2022counterfactual} & $0.98 \pm 0.0$ & $0.95 \pm 0.01$ & $0.86 \pm 0.05$ \\
     \hline
    \multirow{3}{*}{{Law}}
     & EA [Ours] & $1.0 \pm 0.0$ & $1.0 \pm 0.0$ & $1.0 \pm 0.0$ \\
     & Warren et al.~\cite{warren2023explaining} & $0.06 \pm 0.0$ & $0.09 \pm 0.02$ & $0.03 \pm 0.01$ \\
     & Kanamori et al.~\cite{kanamori2022counterfactual} & $0.97 \pm 0.0$ & $0.99 \pm 0.0$ & $0.81 \pm 0.05$ \\
     \hline
\end{tabular}
\end{table}

\begin{table}[t]{}
\caption{Cost (percentage of changed features) of the generated multi-instance counterfactuals. We report the mean and variance (over all folds) rounded to two decimal points -- \emph{smaller} numbers are better.}
\label{tab:exp_results:cost}
\centering
%\small
\begin{tabular}{|c|c||c||c|c|}
    \hline
     Data & Method & No Clustering $\downarrow$ & Clustering $\x_i$ $\downarrow$ & Clustering CFs $\downarrow$ \\
     \hline\hline
     \multirow{3}{*}{{Attr.}}
     & EA [Ours] & $0.73 \pm 0.02$ & $0.69 \pm 0.02$ & $0.55 \pm 0.04$ \\
     & Warren et al.~\cite{warren2023explaining} & $1.0 \pm 0.0$ & $1.0 \pm 0.0$ & $1.0 \pm 0.0$ \\
     & Kanamori et al.~\cite{kanamori2022counterfactual} & $0.02 \pm 0.0$ & $0.06 \pm 0.02$ & $0.05 \pm 0.01$ \\
     \hline
    \multirow{3}{*}{{Law}}
     & EA [Ours] & $0.67 \pm 0.01$ & $0.67 \pm 0.01$ & $0.64 \pm 0.01$ \\
     & Warren et al.~\cite{warren2023explaining} & $1.0 \pm 0.0$ & $1.0 \pm 0.0$ & $1.0 \pm 0.0$ \\
     & Kanamori et al.~\cite{kanamori2022counterfactual} & $0.11 \pm 0.01$ & $0.05 \pm 0.01$ & $0.14 \pm 0.01$ \\
     \hline
\end{tabular}
\end{table}

\section{Conclusion \& Summary}\label{sec:summary}
In this work, we proposed a flexible two-stage algorithm for finding groups of instances for which we can compute cost-efficient multi-instance counterfactual explanations. Our proposed algorithm groups instances so that the single multi-instance counterfactual for each group is as simple as possible (i.e. cost efficient). From the empirical evaluation of the method, we conclude that our proposed algorithm (the grouping and the proposed evolutionary method) has either superior or competitive performance compared to existing methods for computing multi-instance counterfactual explanations.
The main limitation of our method is that it suffers from the necessity of computing single counterfactuals for each instance and this impacts its computational performance. Therefore, as part of our future work, we will investigate how to improve computational performance through the use of approximations of counterfactuals, such as gradients. 

\begin{acknowledgments}
  This research was supported by the Ministry of Culture and Science NRW (Germany) as part of the Lamarr Fellow Network. This publication reflects the views of the authors only.  
\end{acknowledgments}

%%
%% Define the bibliography file to be used
\bibliography{bibliography}

\begin{thebibliography}{26}
\expandafter\ifx\csname natexlab\endcsname\relax\def\natexlab#1{#1}\fi
\providecommand{\url}[1]{\texttt{#1}}
\providecommand{\href}[2]{#2}
\providecommand{\path}[1]{#1}
\providecommand{\DOIprefix}{doi:}
\providecommand{\ArXivprefix}{arXiv:}
\providecommand{\URLprefix}{URL: }
\providecommand{\Pubmedprefix}{pmid:}
\providecommand{\doi}[1]{\href{http://dx.doi.org/#1}{\path{#1}}}
\providecommand{\Pubmed}[1]{\href{pmid:#1}{\path{#1}}}
\providecommand{\bibinfo}[2]{#2}
\ifx\xfnm\relax \def\xfnm[#1]{\unskip,\space#1}\fi
%Type = Article
\bibitem[{Goyal et~al.(2020)Goyal, Knackstedt, Yan, and
  Hassanpour}]{goyal2020artificial}
\bibinfo{author}{M.~Goyal}, \bibinfo{author}{T.~Knackstedt},
  \bibinfo{author}{S.~Yan}, \bibinfo{author}{S.~Hassanpour},
\newblock \bibinfo{title}{Artificial intelligence-based image classification
  methods for diagnosis of skin cancer: Challenges and opportunities},
\newblock \bibinfo{journal}{Computers in biology and medicine}
  \bibinfo{volume}{127} (\bibinfo{year}{2020}) \bibinfo{pages}{104065}.
%Type = Article
\bibitem[{Ferrer et~al.(2021)Ferrer, Van~Nuenen, Such, Cot{\'e}, and
  Criado}]{ferrer2021bias}
\bibinfo{author}{X.~Ferrer}, \bibinfo{author}{T.~Van~Nuenen},
  \bibinfo{author}{J.~M. Such}, \bibinfo{author}{M.~Cot{\'e}},
  \bibinfo{author}{N.~Criado},
\newblock \bibinfo{title}{Bias and discrimination in ai: a cross-disciplinary
  perspective},
\newblock \bibinfo{journal}{IEEE Technology and Society Magazine}
  \bibinfo{volume}{40} (\bibinfo{year}{2021}) \bibinfo{pages}{72--80}.
%Type = Article
\bibitem[{Larsson and Heintz(2020)}]{larsson2020transparency}
\bibinfo{author}{S.~Larsson}, \bibinfo{author}{F.~Heintz},
\newblock \bibinfo{title}{Transparency in artificial intelligence},
\newblock \bibinfo{journal}{Internet Policy Review}  (\bibinfo{year}{2020}).
%Type = Misc
\bibitem[{Commission(2021)}]{euAiAct21}
\bibinfo{author}{E.~Commission}, \bibinfo{title}{Proposal for a regulation
  laying down harmonised rules on artificial intelligence (artificial
  intelligence act) and amending certain union legislative acts},
  \bibinfo{year}{2021}.
%Type = Article
\bibitem[{Dwivedi et~al.(2023)Dwivedi, Dave, Naik, Singhal, Omer, Patel, Qian,
  Wen, Shah, Morgan et~al.}]{dwivedi2023explainable}
\bibinfo{author}{R.~Dwivedi}, \bibinfo{author}{D.~Dave},
  \bibinfo{author}{H.~Naik}, \bibinfo{author}{S.~Singhal},
  \bibinfo{author}{R.~Omer}, \bibinfo{author}{P.~Patel},
  \bibinfo{author}{B.~Qian}, \bibinfo{author}{Z.~Wen},
  \bibinfo{author}{T.~Shah}, \bibinfo{author}{G.~Morgan}, et~al.,
\newblock \bibinfo{title}{Explainable ai (xai): Core ideas, techniques, and
  solutions},
\newblock \bibinfo{journal}{ACM Computing Surveys} \bibinfo{volume}{55}
  (\bibinfo{year}{2023}) \bibinfo{pages}{1--33}.
%Type = Article
\bibitem[{Rawal et~al.(2021)Rawal, Mccoy, Rawat, Sadler, and
  Amant}]{rawal2021recent}
\bibinfo{author}{A.~Rawal}, \bibinfo{author}{J.~Mccoy}, \bibinfo{author}{D.~B.
  Rawat}, \bibinfo{author}{B.~Sadler}, \bibinfo{author}{R.~Amant},
\newblock \bibinfo{title}{Recent advances in trustworthy explainable artificial
  intelligence: Status, challenges and perspectives},
\newblock \bibinfo{journal}{IEEE Transactions on Artificial Intelligence}
  \bibinfo{volume}{1} (\bibinfo{year}{2021}) \bibinfo{pages}{1--1}.
%Type = Article
\bibitem[{Wachter et~al.(2017)Wachter, Mittelstadt, and
  Russell}]{CounterfactualWachter}
\bibinfo{author}{S.~Wachter}, \bibinfo{author}{B.~Mittelstadt},
  \bibinfo{author}{C.~Russell},
\newblock \bibinfo{title}{Counterfactual explanations without opening the black
  box: Automated decisions and the gdpr},
\newblock \bibinfo{journal}{Harv. JL \& Tech.} \bibinfo{volume}{31}
  (\bibinfo{year}{2017}) \bibinfo{pages}{841}.
%Type = Inproceedings
\bibitem[{Byrne(2019)}]{CounterfactualsHumanReasoning}
\bibinfo{author}{R.~M.~J. Byrne},
\newblock \bibinfo{title}{Counterfactuals in explainable artificial
  intelligence (xai): Evidence from human reasoning},
\newblock in: \bibinfo{booktitle}{{IJCAI-19}}, \bibinfo{year}{2019}, pp.
  \bibinfo{pages}{6276--6282}. \DOIprefix\doi{10.24963/IJCAI.2019/876}.
%Type = Inproceedings
\bibitem[{Kanamori et~al.(2022)Kanamori, Takagi, Kobayashi, and
  Ike}]{kanamori2022counterfactual}
\bibinfo{author}{K.~Kanamori}, \bibinfo{author}{T.~Takagi},
  \bibinfo{author}{K.~Kobayashi}, \bibinfo{author}{Y.~Ike},
\newblock \bibinfo{title}{Counterfactual explanation trees: Transparent and
  consistent actionable recourse with decision trees},
\newblock in: \bibinfo{booktitle}{{AISTATS} 2022}, \bibinfo{year}{2022}.
  \URLprefix \url{https://proceedings.mlr.press/v151/kanamori22a.html}.
%Type = Article
\bibitem[{Warren et~al.(2023)Warren, Keane, Gueret, and
  Delaney}]{warren2023explaining}
\bibinfo{author}{G.~Warren}, \bibinfo{author}{M.~T. Keane},
  \bibinfo{author}{C.~Gueret}, \bibinfo{author}{E.~Delaney},
\newblock \bibinfo{title}{Explaining groups of instances counterfactually for
  xai: A use case, algorithm and user study for group-counterfactuals},
\newblock \bibinfo{journal}{arXiv:2303.09297}  (\bibinfo{year}{2023}).
%Type = Misc
\bibitem[{Verma et~al.(2020)Verma, Dickerson, and
  Hines}]{CounterfactualReviewChallenges}
\bibinfo{author}{S.~Verma}, \bibinfo{author}{J.~Dickerson},
  \bibinfo{author}{K.~Hines}, \bibinfo{title}{Counterfactual explanations for
  machine learning: A review}, \bibinfo{year}{2020}.
  \href{http://arxiv.org/abs/2010.10596}{{\tt arXiv:2010.10596}}.
%Type = Inproceedings
\bibitem[{Keane and Smyth(2020)}]{keane2020good}
\bibinfo{author}{M.~T. Keane}, \bibinfo{author}{B.~Smyth},
\newblock \bibinfo{title}{Good counterfactuals and where to find them: A
  case-based technique for generating counterfactuals for explainable ai
  (xai)},
\newblock in: \bibinfo{booktitle}{ICCBR}, \bibinfo{year}{2020}.
%Type = Article
\bibitem[{Guidotti(2022)}]{guidotti2022counterfactual}
\bibinfo{author}{R.~Guidotti},
\newblock \bibinfo{title}{Counterfactual explanations and how to find them:
  literature review and benchmarking},
\newblock \bibinfo{journal}{Data Mining and Knowledge Discovery}
  (\bibinfo{year}{2022}) \bibinfo{pages}{1--55}.
  \DOIprefix\doi{10.1007/s10618-022-00831-6}.
%Type = Inproceedings
\bibitem[{Poyiadzi et~al.(2020)Poyiadzi, Sokol, Santos-Rodriguez, De~Bie, and
  Flach}]{poyiadzi2020face}
\bibinfo{author}{R.~Poyiadzi}, \bibinfo{author}{K.~Sokol},
  \bibinfo{author}{R.~Santos-Rodriguez}, \bibinfo{author}{T.~De~Bie},
  \bibinfo{author}{P.~Flach},
\newblock \bibinfo{title}{Face: Feasible and actionable counterfactual
  explanations},
\newblock \bibinfo{publisher}{Association for Computing Machinery},
  \bibinfo{address}{New York, NY, USA}, \bibinfo{year}{2020}.
  \DOIprefix\doi{10.1145/3375627.3375850}.
%Type = Inproceedings
\bibitem[{Artelt et~al.(2021)Artelt, Vaquet, Velioglu, Hinder, Brinkrolf,
  Schilling, and Hammer}]{artelt2021evaluating}
\bibinfo{author}{A.~Artelt}, \bibinfo{author}{V.~Vaquet},
  \bibinfo{author}{R.~Velioglu}, \bibinfo{author}{F.~Hinder},
  \bibinfo{author}{J.~Brinkrolf}, \bibinfo{author}{M.~Schilling},
  \bibinfo{author}{B.~Hammer},
\newblock \bibinfo{title}{Evaluating robustness of counterfactual
  explanations},
\newblock in: \bibinfo{booktitle}{{IEEE} {SSCI}}, \bibinfo{year}{2021}.
  \DOIprefix\doi{10.1109/SSCI50451.2021.9660058}.
%Type = Article
\bibitem[{Slack et~al.(2021)Slack, Hilgard, Lakkaraju, and
  Singh}]{slack2021counterfactual}
\bibinfo{author}{D.~Slack}, \bibinfo{author}{A.~Hilgard},
  \bibinfo{author}{H.~Lakkaraju}, \bibinfo{author}{S.~Singh},
\newblock \bibinfo{title}{Counterfactual explanations can be manipulated},
\newblock \bibinfo{journal}{Advances in Neural Information Processing Systems}
  \bibinfo{volume}{34} (\bibinfo{year}{2021}) \bibinfo{pages}{62--75}.
%Type = Inproceedings
\bibitem[{Artelt and Hammer(2023)}]{artelt2023ijcai}
\bibinfo{author}{A.~Artelt}, \bibinfo{author}{B.~Hammer},
\newblock \bibinfo{title}{"explain it in the same way!" -- model-agnostic group
  fairness of counterfactual explanations},
\newblock in: \bibinfo{booktitle}{IJCAI Workshop on XAI}, \bibinfo{year}{2023}.
  \URLprefix \url{https://sites.google.com/view/xai2023}.
%Type = Article
\bibitem[{Ley et~al.(2023)Ley, Mishra, and Magazzeni}]{ley2023globe}
\bibinfo{author}{D.~Ley}, \bibinfo{author}{S.~Mishra},
  \bibinfo{author}{D.~Magazzeni},
\newblock \bibinfo{title}{{GLOBE-CE:} {A} translation based approach for global
  counterfactual explanations} \bibinfo{volume}{202} (\bibinfo{year}{2023})
  \bibinfo{pages}{19315--19342}. \URLprefix
  \url{https://proceedings.mlr.press/v202/ley23a.html}.
%Type = Inproceedings
\bibitem[{Artelt and Gregoriades(2023)}]{ArteltG23}
\bibinfo{author}{A.~Artelt}, \bibinfo{author}{A.~Gregoriades},
\newblock \bibinfo{title}{"how to make them stay?": Diverse counterfactual
  explanations of employee attrition},
\newblock in: \bibinfo{booktitle}{{ICEIS}}, \bibinfo{year}{2023}.
  \DOIprefix\doi{10.5220/0011961300003467}.
%Type = Inproceedings
\bibitem[{Rawal and Lakkaraju(2020)}]{DBLP:conf/nips/RawalL20}
\bibinfo{author}{K.~Rawal}, \bibinfo{author}{H.~Lakkaraju},
\newblock \bibinfo{title}{Beyond individualized recourse: Interpretable and
  interactive summaries of actionable recourses},
\newblock in: \bibinfo{booktitle}{NeurIPS}, \bibinfo{year}{2020}. \URLprefix
  \url{https://proceedings.neurips.cc/paper/2020/hash/8ee7730e97c67473a424ccfeff49ab20-Abstract.html}.
%Type = Inproceedings
\bibitem[{Plumb et~al.(2020)Plumb, Terhorst, Sankararaman, and
  Talwalkar}]{plumb2020explaining}
\bibinfo{author}{G.~Plumb}, \bibinfo{author}{J.~Terhorst},
  \bibinfo{author}{S.~Sankararaman}, \bibinfo{author}{A.~Talwalkar},
\newblock \bibinfo{title}{Explaining groups of points in low-dimensional
  representations},
\newblock in: \bibinfo{booktitle}{{ICML} 2020}, volume \bibinfo{volume}{119},
  \bibinfo{publisher}{{PMLR}}, \bibinfo{year}{2020}, pp.
  \bibinfo{pages}{7762--7771}. \URLprefix
  \url{http://proceedings.mlr.press/v119/plumb20a.html}.
%Type = Inproceedings
\bibitem[{Leofante and Potyka(2024)}]{leofante2024promoting}
\bibinfo{author}{F.~Leofante}, \bibinfo{author}{N.~Potyka},
\newblock \bibinfo{title}{Promoting counterfactual robustness through
  diversity},
\newblock in: \bibinfo{booktitle}{Proceedings of the AAAI Conference on
  Artificial Intelligence}, volume~\bibinfo{volume}{38}, \bibinfo{year}{2024},
  pp. \bibinfo{pages}{21322--21330}.
%Type = Inproceedings
\bibitem[{Dominguez-Olmedo et~al.(2022)Dominguez-Olmedo, Karimi, and
  Sch{\"o}lkopf}]{dominguez2022adversarial}
\bibinfo{author}{R.~Dominguez-Olmedo}, \bibinfo{author}{A.~H. Karimi},
  \bibinfo{author}{B.~Sch{\"o}lkopf},
\newblock \bibinfo{title}{On the adversarial robustness of causal algorithmic
  recourse},
\newblock in: \bibinfo{booktitle}{ICML}, \bibinfo{year}{2022}.
%Type = Article
\bibitem[{Reeves(2010)}]{reeves2010genetic}
\bibinfo{author}{C.~R. Reeves},
\newblock \bibinfo{title}{Genetic algorithms},
\newblock \bibinfo{journal}{Handbook of metaheuristics}  (\bibinfo{year}{2010})
  \bibinfo{pages}{109--139}.
%Type = Misc
\bibitem[{IBM(2020)}]{IbmHrAnalyticsEmployee}
\bibinfo{author}{IBM}, \bibinfo{title}{Ibm hr analytics employee},
  \bibinfo{howpublished}{\url{https://www.kaggle.com/pavansubhasht/ibm-hr-analytics-attrition-dataset}},
  \bibinfo{year}{2020}.
%Type = Article
\bibitem[{Wightman(1998)}]{wightman1998lsac}
\bibinfo{author}{L.~F. Wightman},
\newblock \bibinfo{title}{Lsac national longitudinal bar passage study. lsac
  research report series.}  (\bibinfo{year}{1998}).

\end{thebibliography}

%\appendix

\end{document}